\documentclass{article}

\usepackage[final,nonatbib]{neurips_2024}

\usepackage[utf8]{inputenc} 
\usepackage[T1]{fontenc}    
\usepackage[citecolor=blue, colorlinks]{hyperref}
\usepackage{url}            
\usepackage{booktabs}       
\usepackage{amsfonts}       
\usepackage{nicefrac}       
\usepackage{microtype}      
\usepackage{xcolor}         
\usepackage{amsmath}
\usepackage{graphicx}
\usepackage{enumitem}
\usepackage{caption}
\usepackage{multirow}
\usepackage{wrapfig}
\usepackage{pifont}
\usepackage{soul}
\usepackage{fontawesome}
\usepackage{makecell}
\usepackage{threeparttable}
\usepackage{subcaption}

\usepackage{colortbl}
\definecolor{light-gray}{gray}{0.95}

\title{Flatten Anything: \\ Unsupervised Neural Surface Parameterization}

\author{Qijian Zhang\textsuperscript{\rm 1}, Junhui Hou\textsuperscript{\rm 1}\thanks{Corresponding author. This work was supported in part by the National Natural Science Foundation of China Excellent Young Scientists Fund 62422118, and in part by the Hong Kong Research Grants Council under Grants 11219324 and 11219422.}, Wenping Wang\textsuperscript{\rm 2}, Ying He\textsuperscript{\rm 3} \\ 
	\textsuperscript{\rm 1}Department of Computer Science, City University of Hong Kong, Hong Kong SAR, China \\
	\textsuperscript{\rm 2}Department of Computer Science and Engineering, Texas A\&M University, Texas, USA \\
	\textsuperscript{\rm 3}School of Computer Science and Engineering, Nanyang Technological University, Singapore \\
	\texttt{qijizhang3-c@my.cityu.edu.hk, jh.hou@cityu.edu.hk} \\ 
	\texttt{wenping@tamu.edu, yhe@ntu.edu.sg} \\}

\begin{document}

\maketitle

\begin{figure*}[h]
	\centering
	\vspace{-0.5cm}
	\includegraphics[width=1.0\linewidth]{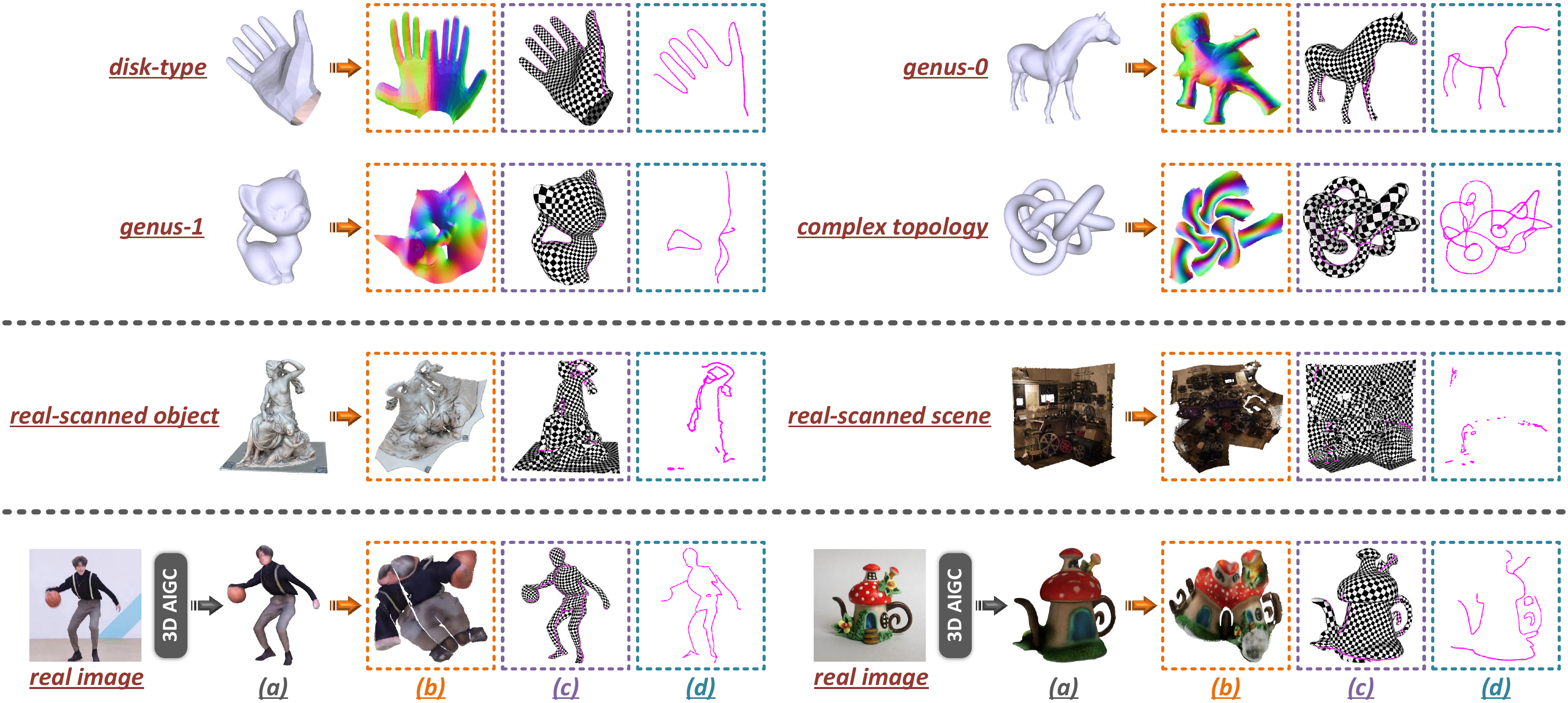}
	\caption{\textbf{\textit{Flatten Anything Model (FAM)}} for neural surface parameterization: (a) input 3D models; (b) learned UV coordinates; (c) texture mappings; (d) learned cutting seams.}
\end{figure*}

\begin{abstract}
	\vspace{-0.3cm}
	Surface parameterization plays an essential role in numerous computer graphics and geometry processing applications. Traditional parameterization approaches are designed for high-quality meshes laboriously created by specialized 3D modelers, thus unable to meet the processing demand for the current explosion of ordinary 3D data. Moreover, their working mechanisms are typically restricted to certain simple topologies, thus relying on cumbersome manual efforts (e.g., surface cutting, part segmentation) for pre-processing. In this paper, we introduce the \textit{Flatten Anything Model (FAM)}, an unsupervised neural architecture to achieve global free-boundary surface parameterization via learning point-wise mappings between 3D points on the target geometric surface and adaptively-deformed UV coordinates within the 2D parameter domain. To mimic the actual physical procedures, we ingeniously construct geometrically-interpretable sub-networks with specific functionalities of surface cutting, UV deforming, unwrapping, and wrapping, which are assembled into a bi-directional cycle mapping framework. Compared with previous methods, our FAM directly operates on discrete surface points without utilizing connectivity information, thus significantly reducing the strict requirements for mesh quality and even applicable to unstructured point cloud data. More importantly, our FAM is fully-automated without the need for pre-cutting and can deal with highly-complex topologies, since its learning process adaptively finds reasonable cutting seams and UV boundaries. Extensive experiments demonstrate the universality, superiority, and inspiring potential of our proposed neural surface parameterization paradigm. Our code is available at \url{https://github.com/keeganhk/FlattenAnything}.
\end{abstract}

\section{Introduction} \label{pp_sec:intro}

Surface parameterization intuitively refers to the process of flattening a 3D geometric surface onto a 2D plane, which is typically called the parameter domain. For any 3D spatial point $(x, y, z)$ lying on the underlying surface, we explicitly map it to a 2D planar coordinate $(u, v)$ while satisfying certain continuity and distortion constraints. Building such point-to-point mappings is also known as UV unwrapping, which serves as an indispensable component in modern graphics rendering pipelines for texture mapping and is also widely used in many downstream geometry processing applications, such as remeshing, mesh completion/compression, detail transfer, surface fitting, and editing, etc.

Theoretically, for any two geometric surfaces with identical/similar topological structures, there exists a bijective mapping between them. Nevertheless, when the topology of the target 3D surface becomes complicated (e.g., with high genus), one must pre-open the original mesh to a sufficiently-developable disk along appropriate cutting seams. Consequently, the current industrial practice for implementing UV unwrapping is typically composed of two stages: (1) manually specifying some necessary cutting seams on the original mesh; (2) applying mature disk-topology-oriented parameterization algorithms (e.g., LSCM~\cite{levy2023least}, ABF~\cite{sheffer2001parameterization,sheffer2005abf++}) which have been well integrated into popular 3D modeling software (e.g., Blender, Unity, 3Ds Max) to produce per-vertex 2D UV coordinates. In practice, such stage-wise UV unwrapping pipeline still shows the following limitations and inconveniences:

\textbf{1}) Commonly-used surface parameterization algorithms are designed for well-behaved meshes, which are typically produced by specialized 3D modelers and technical artists. However, with the advent of user-generated content (UGC) fueled by the rapidly growing 3D sensing, reconstruction~\cite{mildenhall2020nerf,kerbl20233d}, and generation~\cite{nichol2022point,poole2023dreamfusion,lin2023magic3d,siddiqui2023meshgpt} techniques, there emerges an urgent need for dealing with ordinary 3D data possibly with unruly anomalies, inferior triangulations, and non-ideal geometries.

\textbf{2}) Despite the existence of a few early attempts at heuristic seam generation \cite{sheffer2002seamster,erickson2002optimally}, the process of finding high-quality cutting seams in practical applications still relies on manual efforts and personal experience. Hence, the entire workflow remains subjective and semi-automated, leading to reduced reliability and efficiency, especially when dealing with complex 3D models that users are unfamiliar with. Besides, since cutting is actually achieved via edge selection, one may need to repeatedly adjust the distribution of mesh edges to allow the cutting seams to walk through.

\textbf{3}) The procedures of cutting the original mesh into a disk and then flattening the resulting disk onto the parameter domain should have been mutually influenced and jointly optimized; otherwise, the overall surface parameterization results could be sub-optimal.

In recent years, there has emerged a new family of neural parameterization approaches targeted at learning parameterized 3D geometric representations via neural network architectures. The pioneering works of FoldingNet~\cite{yang2018foldingnet} and AtlasNet~\cite{groueix2018papier} are among the best two representatives, which can build point-wise mappings via deforming a pre-defined 2D lattice grid to reconstruct the target 3D shape. RegGeoNet~\cite{zhang2022reggeonet} and Flattening-Net~\cite{zhang2023flattening} tend to achieve fixed-boundary rectangular structurization of irregular 3D point clouds through geometry image \cite{gu2002geometry,losasso2003smooth} representations. However, these two approaches cannot be regarded as real-sense surface parameterization due to their lack of mapping constraints. DiffSR~\cite{bednarik2020shape} and Nuvo~\cite{srinivasan2023nuvo} focus on the local parameterization of the original 3D surface but with explicit and stronger constraints on the learned neural mapping process. Their difference lies in that DiffSR aggregates multi-patch parameterizations to reconstruct the original geometric surface, while Nuvo adaptively assigns surface points to different charts in a probabilistic manner.

In this paper, we make the first attempt to investigate neural surface parameterization featured by both global mapping and free boundary. We introduce the Flatten Anything Model (FAM), a universal and fully-automated UV unwrapping approach to build point-to-point mappings between 3D points lying on the target geometric surface and 2D UV coordinates within the adaptively-deformed parameter domain. In general, FAM is designed as an unsupervised learning pipeline (i.e., no ground-truth UV maps are collected), running in a per-model overfitting manner. To mimic the actual physical process, we ingeniously design a series of geometrically-meaningful sub-networks with specific functionalities of surface cutting, UV deforming, 3D-to-2D unwrapping, and 2D-to-3D wrapping, which are further assembled into a bi-directional cycle mapping framework. By optimizing a combination of different loss functions and auxiliary differential geometric constraints, FAM is able to find reasonable 3D cutting seams and 2D UV boundaries, leading to superior parameterization quality when dealing with different degrees of geometric and topological complexities. Comprehensive experiments demonstrate the advantages of our approach over traditional state-of-the-art approaches. Conclusively, our FAM shows the following several major aspects of characteristics and superiorities:
\begin{itemize}[leftmargin=2em,itemsep=2pt,parsep=1pt,topsep=0pt,partopsep=1pt]
	
	\item Compared with traditional mesh parameterization approaches \cite{levy2023least,sheffer2005abf++,rabinovich2017scalable}, FAM directly operates on discrete surface points and jointly learns surface cutting, which can be insensitive to mesh triangulations and is able to deal with non-ideal geometries and arbitrarily-complex topologies. Besides, such a pure neural learning architecture can naturally exploit the powerful parallelism of GPUs and is much easier for implementing, tuning, and further extension.
	
	\item Another distinct advantage lies in the smoothness of parameterization. Since mesh parameterization computes parameters solely for existing vertices, for any point on the mesh that is not a vertex, these approaches resort to interpolation within its encompassing triangle to determine the corresponding UV coordinate. Such practice fails to guarantee smoothness, particularly across edges, and this issue is further exacerbated in low-resolution meshes. By contrast, FAM exploits the inherent smooth property \cite{rahaman2019spectral} of neural networks to learn arbitrary point-to-point mappings, thereby ensuring smoothness for all points on the target surface.
	
	\item Different from previous multi-patch local parameterization frameworks \cite{groueix2018papier,bednarik2020shape,srinivasan2023nuvo}, FAM focuses on global parameterization, a more valuable yet much harder problem setting.
	
	\item  Different from previous fixed-boundary structurization frameworks \cite{zhang2022reggeonet,zhang2023flattening}, FAM deforms the 2D UV parameter domain adaptively and regularizes the learned neural mapping explicitly, thus significantly reducing discontinuities and distortions.
	
\end{itemize}
\vspace{-0.2cm}

\section{Related Works} \label{pp_sec:rw}

\subsection{Traditional Parameterization Approaches} \label{pp_sec:rw-traditional-para}
\vspace{-0.1cm}

Early approaches formulate mesh parameterization as a Laplacian problem, where boundary points are anchored to a pre-determined convex 2D curve~\cite{10.1145/218380.218440,DBLP:journals/cagd/Floater97a}, tackled by sparse linear system solvers. Such a linear strategy is appreciated for its simplicity, efficiency, and the guarantee of bijectivity. However, the rigidity of fixing boundary points in the parameter domain usually leads to significant distortions. In response to these issues, free-boundary parameterization algorithms~\cite{sheffer2005abf++,DBLP:journals/cgf/LiuZXGG08} have been proposed, offering a more flexible setting by relaxing boundary constraints. Although these approaches bring certain improvements, they often struggle to maintain global bijectivity. More recent state-of-the-art approaches~\cite{rabinovich2017scalable,DBLP:journals/tog/SmithS15} shift towards minimizing simpler proxy energies by alternating between local and global optimization phases, which not only accelerate convergence but also enhance the overall parameterization quality. OptCuts~\cite{li2018optcuts} explores joint optimization of surface cutting and mapping distortion. Despite the continuous advancements, the working mechanisms of all these approaches fundamentally rely on the connectivity information inherent in mesh structures, casting doubt on their applicability to unstructured point clouds. Moreover, mesh-oriented parameterization approaches typically assume that the inputs are well-behaved meshes possessing relatively regular triangulations. When faced with input meshes of inferior quality characterized by irregular triangulations and anomalies, remeshing becomes an indispensable procedure. Additionally, how to deal with non-ideal geometries and complex topologies (e.g., non-manifolds, multiply-connected components, thin structures) has always been highly challenging.

Unlike surface meshes, unstructured point clouds lack information regarding the connectivity between points, and thus are much more difficult to parameterize. Hence, only a relatively limited amount of prior works~\cite{zhu2022review} have focused on the parameterization of surface point clouds. Early studies investigate various specialized parameterization strategies for disk-type \cite{floater2001meshless,azariadis2007product,zhang2010mesh}, genus-0 \cite{zwicker2004meshing,liang2012geometric}, genus-1 \cite{tewari2006meshing}, well-sampled \cite{guo2006meshless} or low-quality \cite{meng2013parameterization} point clouds. Later approaches achieve fixed-boundary spherical/rectangular parameterization through approximating the Laplace-Beltrami operators \cite{choi2016spherical} or Teichm\"uller extremal mappings \cite{meng2016tempo}. More recently, FBCP-PC~\cite{choi2022free} further pursues free-boundary conformal parameterization, in which a new point cloud Laplacian approximation scheme is proposed for handling boundary non-convexity.

\subsection{Neural Parameterization Approaches} \label{pp_sec:rw-neural-para}
\vspace{-0.1cm}

Driven by the remarkable success of deep learning, there is a family of recent works applying neural networks to learn parameterized 3D geometric representations. FoldingNet~\cite{yang2018foldingnet} proposes to deform a uniform 2D grid to reconstruct the target 3D point cloud for unsupervised geometric feature learning. AtlasNet~\cite{groueix2018papier} applies multi-grid deformation to learn locally parameterized representations. Subsequently, a series of follow-up researches inherit such a ``folding-style'' parameterization paradigm as pioneered by \cite{yang2018foldingnet,groueix2018papier} and investigate different aspects of modifications. GTIF~\cite{chen2019deep} introduces graph topology inference and filtering mechanisms, empowering the decoder to preserve more representative geometric features in the latent space. EleStruc~\cite{deprelle2019learning} proposes to perform shape reconstruction from learnable 3D elementary structures, rather than a pre-defined 2D lattice. Similarly, TearingNet~\cite{pang2021tearingnet} adaptively breaks the edges of an initial primitive graph for emulating the topology of the target 3D point cloud, which can effectively deal with higher-genus or multi-object inputs. In fact, the ultimate goal of the above line of approaches is to learn expressive shape codewords by means of deformation-driven 3D surface reconstruction. The characteristics of surface parameterization, i.e., the mapping process between 3D surfaces and 2D parameter domains, are barely considered.

In contrast to the extensive research in the fields of deep learning-based 3D geometric reconstruction and feature learning, there only exist a few studies that particularly focus on neural surface parameterization. DiffSR~\cite{bednarik2020shape} adopts the basic multi-patch reconstruction framework \cite{groueix2018papier} and explicitly regularizes multiple differential surface properties. NSM~\cite{morreale2021neural} explores neural encoding of surface maps by overfitting a neural network to an existing UV parameterization pre-computed via standard mesh parameterization algorithms \cite{tutte1963draw,rabinovich2017scalable}. DA-Wand~\cite{liu2023wand} constructs a parameterization-oriented mesh segmentation framework. Around a specified triangle, it learns to select a local sub-region, which is supposed to be sufficiently developable to produce low-distortion parameterization. Inheriting the geometry image (GI) \cite{gu2002geometry} representation paradigm, RegGeoNet~\cite{zhang2022reggeonet} and Flattening-Net~\cite{zhang2023flattening} propose to learn deep regular representations of unstructured 3D point clouds. However, these two approaches lack explicit constraints on the parameterization distortions, and their local-to-global assembling procedures are hard to control. More recently, Nuvo~\cite{srinivasan2023nuvo} proposes a neural UV mapping framework that operates on oriented 3D points sampled from arbitrary 3D representations, liberating from the stringent quality demands of mesh triangulation. This approach assigns the original surface to multiple charts and ignores the packing procedure, thus essentially differing from our targeted global parameterization setting.

\section{Proposed Method} \label{pp_sec:proposed-method}

Our FAM operates on discrete spatial points lying on the target 3D geometric surface for achieving global free-boundary parameterization in an unsupervised learning manner. Denote by $\mathbf{P} \in \mathbb{R}^{N \times 3}$ a set of unstructured points without any connectivity information, we aim at point-wisely parameterizing (i.e., row-wisely mapping) $\mathbf{P}$ onto the planar domain to produce UV coordinates $\mathbf{Q} \in \mathbb{R}^{N \times 2}$.

\textbf{Technical Motivation.} The most straightforward way of learning such a surface flattening process is to directly impose certain appropriate regularizers over the resulting 2D UV coordinates $\mathbf{Q}$. \textit{However, due to the missing of ground-truths, it is impossible to explicitly supervise the generation of $\mathbf{Q}$, and it empirically turns out that designing and minimizing such regularizers cannot produce satisfactory results.} Alternatively, another feasible scheme is to choose an opposite mapping direction to wrap the adaptively-deformed and potentially-optimal 2D UV coordinates $\mathbf{\hat{Q}} \in \mathbb{R}^{N \times 2}$ onto the target surface by generating a new set of unstructured points $\mathbf{\hat{P}} \in \mathbb{R}^{N \times 3}$, which should approximate the input $\mathbf{P}$ by minimizing certain point set similarity metrics. Under ideal situations where $\mathbf{\hat{P}}$ and $\mathbf{P}$ are losslessly matched, we can equivalently deduce the desired UV coordinate of each point in $\mathbf{P}$ (i.e., for a certain 3D point $\mathbf{p}_i \in \mathbf{P}$ perfectly matched with a generated 3D point $\mathbf{\hat{p}}_j \in \mathbf{\hat{P}}$, which is row-wisely mapped from $\mathbf{\hat{q}}_j \in \mathbf{\hat{Q}}$, then we know the UV coordinate of $\mathbf{p}_i$ should be $\mathbf{\hat{q}}_j$). \textit{However, due to the practical difficulty of reconstructing point sets with high-precision point-wise matching, the generated $\mathbf{\hat{P}}$ is only able to coarsely recover the target surface, thereby $\mathbf{\hat{Q}}$ cannot be treated as the resulting 2D UV coordinates of the input $\mathbf{P}$.} Still, despite the inaccuracy of such inverse 2D-to-3D wrapping process, it provides valuable cues indicating how the target 3D surface is roughly transformed from the 2D parameter domain, which naturally motivates us to combine the two opposite mapping processes into a bi-directional joint learning architecture. \textit{To build associations between the unwrapping and wrapping processes and promote mapping bijectivity, we further introduce cycle mapping mechanisms and impose consistency constraints}. Thus, our FAM is designed as a bi-directional cycle mapping workflow, in which all sub-networks are \textit{parameter-sharing} and \textit{jointly-optimized}.

\begin{figure*}[t!]
	\centering
	\includegraphics[width=1.0\linewidth]{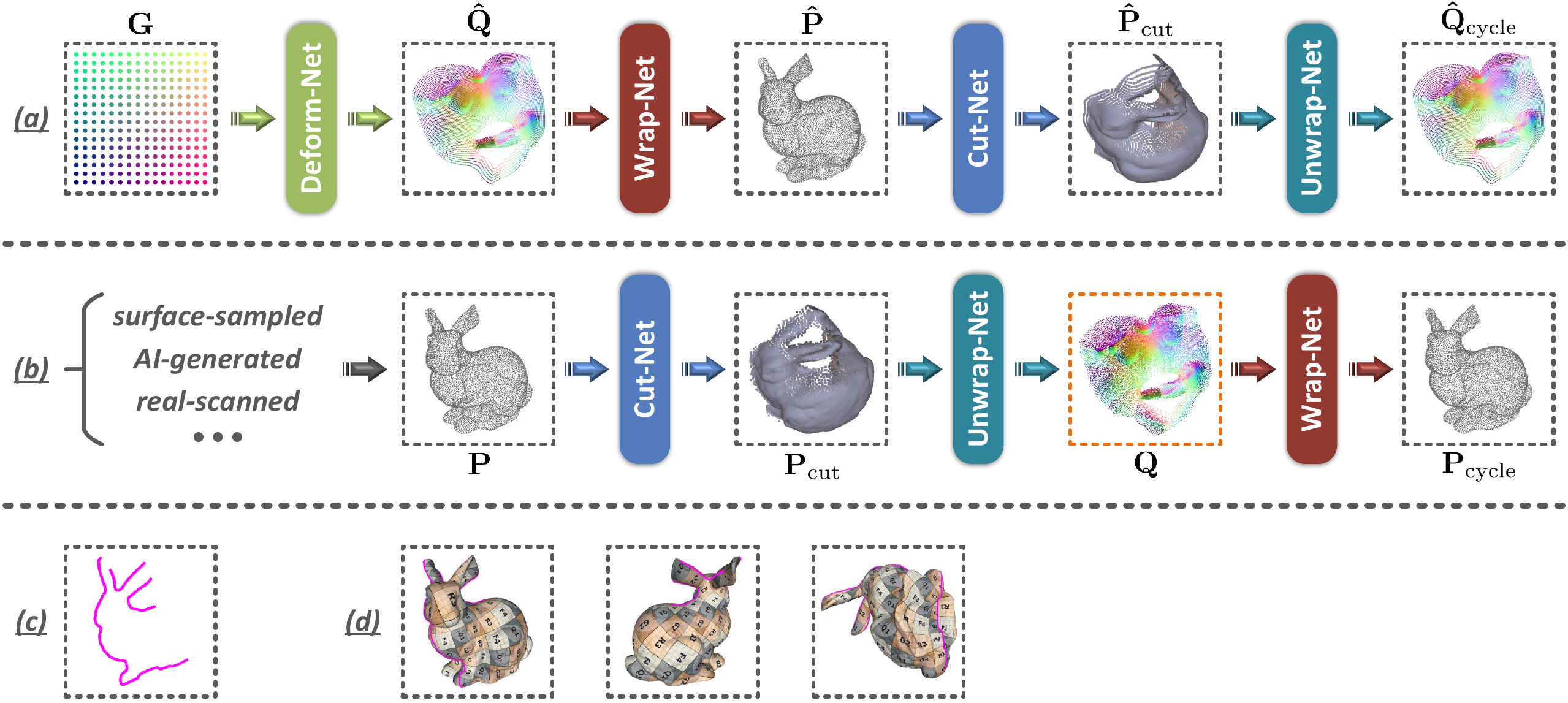}
	\caption{Illustration of bi-directional cycle mapping, composed of (a) 2D$\rightarrow$3D$\rightarrow$2D cycle mapping branch, and (b) 3D$\rightarrow$2D$\rightarrow$3D cycle mapping branch. Modules with the same color share network parameters. (c) shows the learned cutting seams. (d) shows the checker-image texture mapping.}
	\label{fig:overall_workflow}
	\vspace{-0.2cm}
\end{figure*}

\subsection{Bi-directional Cycle Mapping} \label{pp_sec:method-bidir-cycle}

As illustrated in Figure~\ref{fig:overall_workflow}, there are two parallel learning branches comprising a series of geometrically-meaningful sub-networks built upon point-wisely shared multi-layer perceptrons (MLPs):

\begin{itemize}[leftmargin=2em, itemsep=1pt,parsep=1pt,topsep=0pt,partopsep=0pt]
	
	\item The deforming network $\mathcal{M}_d$ (\textbf{Deform-Net}) takes a uniform 2D lattice as input, and deforms the initial grid points to potentially-optimal UV coordinates.
	
	\item The wrapping network $\mathcal{M}_w$ (\textbf{Wrap-Net}) wraps the potentially-optimal 2D UV coordinates onto the target 3D geometric surface.
	
	\item The surface cutting network $\mathcal{M}_c$ (\textbf{Cut-Net}) finds appropriate cutting seams for transforming the original 3D geometric structure to an open and highly-developable surface manifold.
	
	\item The unwrapping network $\mathcal{M}_u$ (\textbf{Unwrap-Net}) assumes that its input surface has already been pre-opened, and flattens the 3D points onto the 2D parameter domain.
	
\end{itemize}


\subsubsection{2D$\rightarrow$3D$\rightarrow$2D Cycle Mapping}

The upper learning branch begins with a pre-defined 2D lattice with $N$ grid points uniformly sampled within $[-1, 1]^2$, denoted as $\mathbf{G} \in \mathbb{R}^{N \times 2}$. To perform adaptive UV space deformation, we employ an offset-based coordinate updating strategy by feeding $\mathbf{G}$ into the Deform-Net to produce another set of 2D UV coordinates $\mathbf{\hat{Q}} \in \mathbb{R}^{N \times 2}$, which can be formulated as:
\begin{equation}
	\mathbf{\hat{Q}} = \mathcal{M}_d(\mathbf{G}) = \xi_d^{\prime\prime}([\xi_d^{\prime}(\mathbf{G}); \mathbf{G}]) + \mathbf{G},
\end{equation}
\noindent where $[\ast;\ast]$ denotes channel concatenation, $\xi_d^{\prime}: \mathbb{R}^2 \rightarrow \mathbb{R}^h$ and $\xi_d^{\prime\prime}: \mathbb{R}^{(h+2)} \rightarrow \mathbb{R}^2$ are stacked MLPs. Intuitively, the initial 2D grid points are first embedded through $\xi_d^{\prime}$ into the $h$-dimensional latent space, concatenated with itself, and then mapped onto the 2D planar domain through $\xi_d^{\prime\prime}$. The learned offsets are point-wisely added to the initial grid points to produce the resulting $\mathbf{\hat{Q}}$.

After that, we perform 2D-to-3D wrapping by feeding the generated $\mathbf{\hat{Q}}$ into the Wrap-Net to produce $\mathbf{\hat{P}} \in \mathbb{R}^{N \times 3}$, which is expected to roughly approximate the target 3D geometric structure represented by the input 3D point set $\mathbf{P}$. In the meantime, we use three more output channels to produce normals $\mathbf{\hat{P}^n} \in \mathbb{R}^{N \times 3}$. The whole network behavior can be described as:
\begin{equation}
	[\mathbf{\hat{P}}; \mathbf{\hat{P}^n}] = \mathcal{M}_w(\mathbf{\hat{Q}}) = \xi_w^{\prime\prime}([\xi_w^{\prime}(\mathbf{\hat{Q}}); \mathbf{\hat{Q}}]),
\end{equation}
\noindent where $\xi_w^{\prime}: \mathbb{R}^2 \rightarrow \mathbb{R}^h$ and $\xi_w^{\prime\prime}: \mathbb{R}^{(h+2)} \rightarrow \mathbb{R}^6$ are stacked MLPs, and the resulting $6$ output channels are respectively split into $\mathbf{\hat{P}}$ and $\mathbf{\hat{P}^n}$.

In order to construct cycle mapping, we further apply the Cut-Net on the reconstructed $\mathbf{\hat{P}}$, which is transformed into an open 3D surface manifold $\mathbf{\hat{P}}_\mathrm{cut} \in \mathbb{R}^{N \times 3}$, as given by:
\begin{equation}
	\mathbf{\hat{P}}_\mathrm{cut} = \mathcal{M}_c(\mathbf{\hat{P}}) = \xi_c^{\prime\prime}([\xi_c^{\prime}(\mathbf{\hat{P}}); \mathbf{\hat{P}}]) + \mathbf{\hat{P}},
\end{equation}
\noindent where $\xi_c^{\prime}: \mathbb{R}^3 \rightarrow \mathbb{R}^h$ and $\xi_c^{\prime\prime}: \mathbb{R}^{(h+3)} \rightarrow \mathbb{R}^3$ are stacked MLPs.

Once we obtain the highly-developable 3D surface manifold $\mathbf{\hat{P}}_\mathrm{cut}$, it is straightforward to perform 3D-to-2D flattening through the Unwrap-Net, which can be formulated as:
\begin{equation}
	\mathbf{\hat{Q}}_\mathrm{cycle} = \mathcal{M}_u(\mathbf{\hat{P}}_\mathrm{cut}) = \xi_u(\mathbf{\hat{P}}_\mathrm{cut}),
\end{equation}
\noindent where $\xi_u: \mathbb{R}^3 \rightarrow \mathbb{R}^2$ is simply implemented as stacked MLPs, and the resulting $\mathbf{\hat{Q}}_\mathrm{cycle} \in \mathbb{R}^{N \times 2}$ is expected to be row-wisely equal to $\mathbf{\hat{Q}}$.

\subsubsection{3D$\rightarrow$2D$\rightarrow$3D Cycle Mapping}

The bottom learning branch directly starts by feeding the input 3D point set $\mathbf{P}$ into the Cut-Net to produce an open 3D surface manifold $\mathbf{P}_\mathrm{cut}$, which can be formulated as:
\begin{equation}
	\mathbf{P}_\mathrm{cut}  = \mathcal{M}_c(\mathbf{P}) = \xi_c^{\prime\prime}([\xi_c^{\prime}(\mathbf{P}); \mathbf{P}]) + \mathbf{P}.
\end{equation}

After that, we apply the Unwrap-Net on the generated $\mathbf{P}_\mathrm{cut}$ to produce the desired 2D UV coordinates $\mathbf{Q}$, as given by:
\begin{equation}
	\mathbf{Q}  = \mathcal{M}_u(\mathbf{P}_\mathrm{cut}) = \xi_u(\mathbf{P}_\mathrm{cut}).
\end{equation}

In order to construct cycle mapping, we further apply the Wrap-Net on the generated $\mathbf{Q}$ for recovering the target 3D geometric structure, which can be formulated as:
\begin{equation}
	[\mathbf{P}_\mathrm{cycle}; \mathbf{P}_\mathrm{cycle}^\mathbf{n}]  = \mathcal{M}_w(\mathbf{Q}) = \xi_w^{\prime\prime}([\xi_w^{\prime}(\mathbf{Q}); \mathbf{Q}]),
\end{equation}
\noindent where $\mathbf{P}_\mathrm{cycle}$ is expected to be row-wisely equal to $\mathbf{P}$. When the input surface points are accompanied by normals $\mathbf{P}^\mathbf{n} \in \mathbb{R}^{N \times 3}$, we also expect that the normal directions of $\mathbf{P}^\mathbf{n}$ and the generated $\mathbf{P}_\mathrm{cycle}^\mathbf{n}$ should be row-wisely consistent.

\textbf{Remark.} Throughout the overall bi-directional cycle mapping framework, \textit{only $\mathbf{Q}$ is regarded as the desired 2D UV coordinates that are row-wisely mapped from the input 3D surface points $\mathbf{P}$}, all the others (e.g., $\mathbf{\hat{Q}}$ and $\mathbf{\hat{Q}}_\mathrm{cycle}$) just serve as intermediate results. Morevoer, it is also worth emphasizing that, \textit{when the training is finished, any arbitrary surface-sampled 3D points can be fed into our FAM to obtain point-wise UV coordinates}.

\textbf{Extraction of Cutting Seams.} The extraction of points located on the learned cutting seams, which we denote as $\mathbf{E} = \{ \mathbf{e}_i \in \mathbf{P} \}$, can be conveniently achieved by comparing the mapping relationships between the input $\mathbf{P}$ and the parameterized $\mathbf{Q}$.

Specifically, since $\mathbf{P}$ and $\mathbf{Q}$ are mapped in a row-wise manner, for each input 3D point $\mathbf{p}_i \in \mathbf{P}$ and its K-nearest neighbors $\{ \mathbf{p}_i^{(k)} \}_{k=1}^{K_\mathrm{cut}}$ within $\mathbf{P}$, we can directly know their 2D UV coordinates, denoted as $\mathbf{q}_i \in \mathbf{Q}$ and $\{ \mathbf{q}_i^{(k)} \}_{k=1}^{K_\mathrm{cut}}$. The maximum distance value $\eta_i$ between $\mathbf{q}_i$ and its neighboring points can be computed as:
\begin{equation} \label{eqn:find-cut}
	\eta_i = \mathrm{max}(\{ \lVert \mathbf{q}_i - \mathbf{q}_i^{(k)} \rVert_2 \}_{k=1}^{K_\mathrm{cut}}).
\end{equation}
Then, $\mathbf{p}_i$ will be determined to locate on the cutting seam if $\eta_i$ is larger than a certain threshold $T_\mathrm{cut}$. Collecting all such points within $\mathbf{P}$ can deduce a set of seam points $\mathbf{E}$, and it is observed that in this way no seam points would be identified if cutting the original 3D surface is actually unnecessary.

\subsection{Training Objectives} \label{pp_sec:training-objectives}

As an unsupervised learning framework, FAM is trained by minimizing a series of carefully-designed objective functions and constraints, whose formulations and functionalities are introduced below.

\noindent \textbf{Unwrapping Loss.} The most fundamental requirement for $\mathbf{Q}$ is that any two parameterized coordinates cannot overlap. Hence, for each 2D UV coordinate $\mathbf{q}_i$, we search its K-nearest-neighbors $\{ \mathbf{q}_{i}^{(k)} \}_{k=1}^{K_u}$, and penalize neighboring points that are too close, as given by:
\begin{equation} \label{eqn:unwrap-loss}
	\ell_\mathrm{unwrap} = \sum_{i=1}^{N} \sum_{k=1}^{K_\mathrm{u}} \mathrm{max}(0, \epsilon - \lVert \mathbf{q}_i - \mathbf{q}_{i}^{(k)} \rVert_2),
\end{equation}
\noindent where the minimum distance value allowed between neighboring points is controlled by a threshold $\epsilon$. \\

\noindent \textbf{Wrapping Loss.} The major supervision of the upper learning branch is that the generated point set $\hat{\mathbf{P}}$ should roughly approximate the original 3D geometric structure represented by $\mathbf{P}$. Since $\hat{\mathbf{P}}$ and $\mathbf{P}$ are not row-wisely corresponded, we use the commonly-used Chamfer distance $\texttt{CD}(\ast;\ast)$ to measure point set similarity, as given by:
\begin{equation}
	\ell_\mathrm{wrap} = \texttt{CD}(\hat{\mathbf{P}};\mathbf{P}).
\end{equation}
Here, there is no need to impose the unwrapping loss over $\hat{\mathbf{P}}$, since we empirically find that optimizing the CD loss naturally suppresses clustered point distribution. Besides, we do not optimize $\texttt{CD}(\hat{\mathbf{Q}};\mathbf{Q})$, since we observe degraded performances, possibly because in earlier training iterations $\hat{\mathbf{Q}}$ and $\mathbf{Q}$ are far from optimal and thus can interfere with each other.

\noindent \textbf{Cycle Consistency Loss.} Our bi-directional cycle mapping framework aims to promote consistencies within both 3D and 2D domains, which are natural to learn thanks to our symmetric designs of the functionalities of sub-networks. Thus, the cycle consistency loss is formulated as:
\begin{equation} \label{eqn:cc-loss}
	\ell_\mathrm{cycle} = \lVert \mathbf{P} - \mathbf{P}_\mathrm{cycle} \rVert_1 + 
	\lVert \hat{\mathbf{Q}} - \hat{\mathbf{Q}}_\mathrm{cycle} \rVert_1 + 
	\texttt{CS}(\mathbf{P}^\mathbf{n}; \mathbf{P}_\mathrm{cycle}^\mathbf{n}),
\end{equation}
\noindent where $\texttt{CS}(\ast;\ast)$ computes point-wise cosine similarity between $\mathbf{P}_\mathrm{cycle}^\mathbf{n}$ and ground-truth normals $\mathbf{P}^\mathbf{n}$, which can be removed when ground-truth normals are not provided or hard to compute.

\noindent \textbf{Mapping Distortion Constraint.} We regularize the distortion of the learned neural mapping function by exploiting differential surface properties, which can be conveniently deduced via the automatic differentiation mechanisms in common deep learning programming frameworks.

Throughout the bi-directional cycle mapping framework, 2D UV coordinates $\mathbf{Q}$ and $\hat{\mathbf{Q}}$ are respectively mapped to 3D surface points $\mathbf{P}_\mathrm{cycle}$ and $\hat{\mathbf{P}}$. For convenience, here we denote the two neural mapping functions as $f: \mathbb{R}^2 \rightarrow \mathbb{R}^3$ and $g: \mathbb{R}^2 \rightarrow \mathbb{R}^3$. We compute the derivatives of $f$ and $g$ with respect to $\mathbf{Q}$ and $\hat{\mathbf{Q}}$ at each 2D UV point $(u, v)$ to obtain the Jacobian matrices $\mathbf{J}_f \in \mathbb{R}^{3 \times 2}$ and $\mathbf{J}_g \in \mathbb{R}^{3 \times 2}$:
\begin{equation}
	\mathbf{J}_f = (f_u \enspace f_v), ~ \mathbf{J}_g = (g_u \enspace g_v)
\end{equation}
\noindent where $f_u$, $f_v$, $g_u$, $g_v$ are three-dimensional vectors of partial derivatives. Then we further deduce the eigenvalues of $\mathbf{J}_f^T \mathbf{J}_f$ and  $\mathbf{J}_g^T\mathbf{J}_g$, which are denoted as $(\lambda_f^1, \lambda_f^2)$ and $(\lambda_g^1, \lambda_g^2)$. Thus, we can promote conformal (i.e., angle-preserving) parameterizations by constraining the following regularizer:
\begin{equation}
	\ell_\mathrm{conf} = \\
	\sum_{\mathbf{q} \in \mathbf{Q}} \lVert \lambda_f^1 - \lambda_f^2 \rVert_1 + \\ 
	\sum_{\hat{\mathbf{q}} \in \hat{\mathbf{Q}}}\lVert \lambda_g^1 - \lambda_g^2 \rVert_1.
\end{equation}

\section{Experiments} \label{pp_sec:experiments}

We collected a series of 3D surface models with different types and complexities of geometric and/or topological structures for experimental evaluation. We made qualitative and quantitative comparisons with SLIM~\cite{rabinovich2017scalable}, a state-of-the-art and widely-used mesh parameterization algorithm. Additionally, we also compared our FAM with FBCP-PC~\cite{choi2022free} for unstructured and unoriented (i.e., without normals) 3D point cloud parameterization. Finally, we conducted ablation studies to demonstrate the effectiveness of the proposed bi-directional cycle mapping framework and the necessity of Cut-Net, together with different aspects of verifications for comprehensively understanding the characteristics of our approach. Due to page limitations, detailed technical implementations and additional experimental results are presented in our Appendix

\begin{figure*}[t!]
	\centering
	\includegraphics[width=1.0\linewidth]{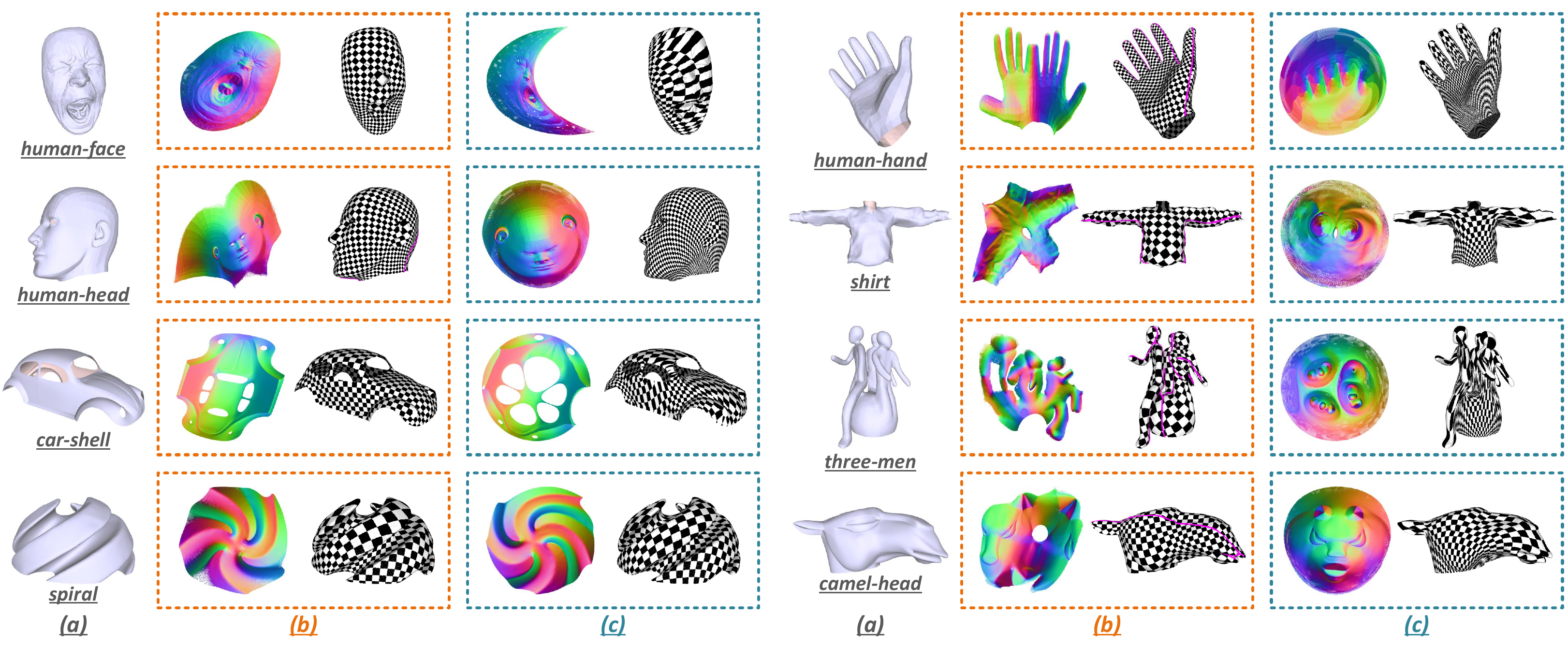}
	\caption{Comparison of UV unwrapping and texture mapping results on different (a) open surface models produced by (b) our FAM and (c) SLIM, where the 2D UV coordinates are color-coded by ground-truth point-wise normals to facilitate visualization.}
	\label{fig:comparison_with_slim}
	\vspace{-0.4cm}
\end{figure*}

\noindent \textbf{Comparison with SLIM~\cite{rabinovich2017scalable}.} Considering that SLIM is only able to be applied on disk-topologies or surfaces with open boundaries, we collected $8$ such testing models for parameterization, using its officially-released code. For the training of our FAM, we uniformly sample $10,000$ points from the original mesh vertices at each optimization iteration. During testing, the whole vertex set is fed into our FAM to produce per-vertex UV coordinates. Then, we can perform checker-map texture mappings for qualitative evaluation. For quantitative evaluation, we computed conformality metrics (\textit{the lower, the better}) by measuring the average of the absolute angle differences between each corresponding angle of the 3D triangles and the 2D parameterized triangles. As shown in Figure~\ref{fig:comparison_with_slim} and Table~\ref{tab:quant_comparison_with_slim}, our FAM outperforms SLIM both qualitatively and quantitatively on all testing models.

\begin{table}[t!]
	\centering	
	\renewcommand\arraystretch{1.0}
	\setlength{\tabcolsep}{3.0pt}
	\caption{Quantitative comparisons of our FAM and SLIM in terms of parameterization conformality.}
	\begin{tabular}{ c | c | c | c | c | c | c | c | c } 
		\toprule[1.0pt]
		Model & \textit{human-face} & \textit{human-head} & \textit{car-shell} & \textit{spiral} & \textit{human-hand} & \textit{shirt} & \textit{three-men} & \textit{camel-head} \\
		\hline\hline
		SLIM & 0.635 & 0.254 & 0.411 & 0.114 & 0.609 & 0.443 & 0.645 & 0.349 \\
		\hline
		\rowcolor{light-gray}
		FAM & \textbf{0.074} & \textbf{0.094} & \textbf{0.037} & \textbf{0.087} & \textbf{0.145} & \textbf{0.166} & \textbf{0.162} & \textbf{0.088} \\
		\bottomrule[1.0pt]
	\end{tabular}
	\label{tab:quant_comparison_with_slim}
	\vspace{-0.4cm}
\end{table}

\begin{table}[h!]
	\centering	
	\renewcommand\arraystretch{1.0}
	\setlength{\tabcolsep}{3.88pt}
	\caption{Quantitative conformality metrics of our parameterization results.}
	\begin{tabular}{ c | c | c | c | c | c | c | c} 
		\toprule[1.0pt]
		Model & \textit{brain} & \textit{cow} & \textit{fandisk} & \textit{human-body} & \textit{lion} & \textit{nail} & \textit{nefertiti} \\
		\hline
		\rowcolor{light-gray}
		Conf. Metric & 0.263 & 0.210 & 0.105 & 0.198 & 0.192 & 0.162 & 0.117 \\
		\hline\hline
		Model & \textit{bucket} & \textit{dragon} & \textit{mobius-strip} & \textit{rocker-arm} & \textit{torus-double} & \textit{three-holes} & \textit{fertility} \\
		\hline
		\rowcolor{light-gray}
		Conf. Metric & 0.179 & 0.311 & 0.062 & 0.143 & 0.126 & 0.142 & 0.166 \\
		\bottomrule[1.0pt]
	\end{tabular}
	\label{tab:quant_ours}
	\vspace{-0.3cm}
\end{table}

\begin{table}[h!]
	\centering	
	\renewcommand\arraystretch{1.0}
	\setlength{\tabcolsep}{8.0pt}
	\caption{Conformality metrics of our FAM and FBCP-PC for point cloud parameterization.}
	\begin{tabular}{ c | c | c | c } 
		\toprule[1.0pt]
		Model & \textit{cloth-pts (\#Pts=7K)}  & \textit{julius-pts (\#Pts=11K)}  & \textit{spiral-pts (\#Pts=28K)} \\
		\hline\hline
		FBCP-PC & \textbf{0.021} & \textbf{0.019} & \textbf{0.023} \\
		\hline
		\rowcolor{light-gray}
		FAM & 0.037 & 0.058 & 0.117 \\
		\bottomrule[1.0pt]
	\end{tabular}
	\label{tab:quant_point_para_comparison}
	\vspace{-0.4cm}
\end{table}

\begin{figure*}[t!]
	\centering
	\includegraphics[width=0.90\linewidth]{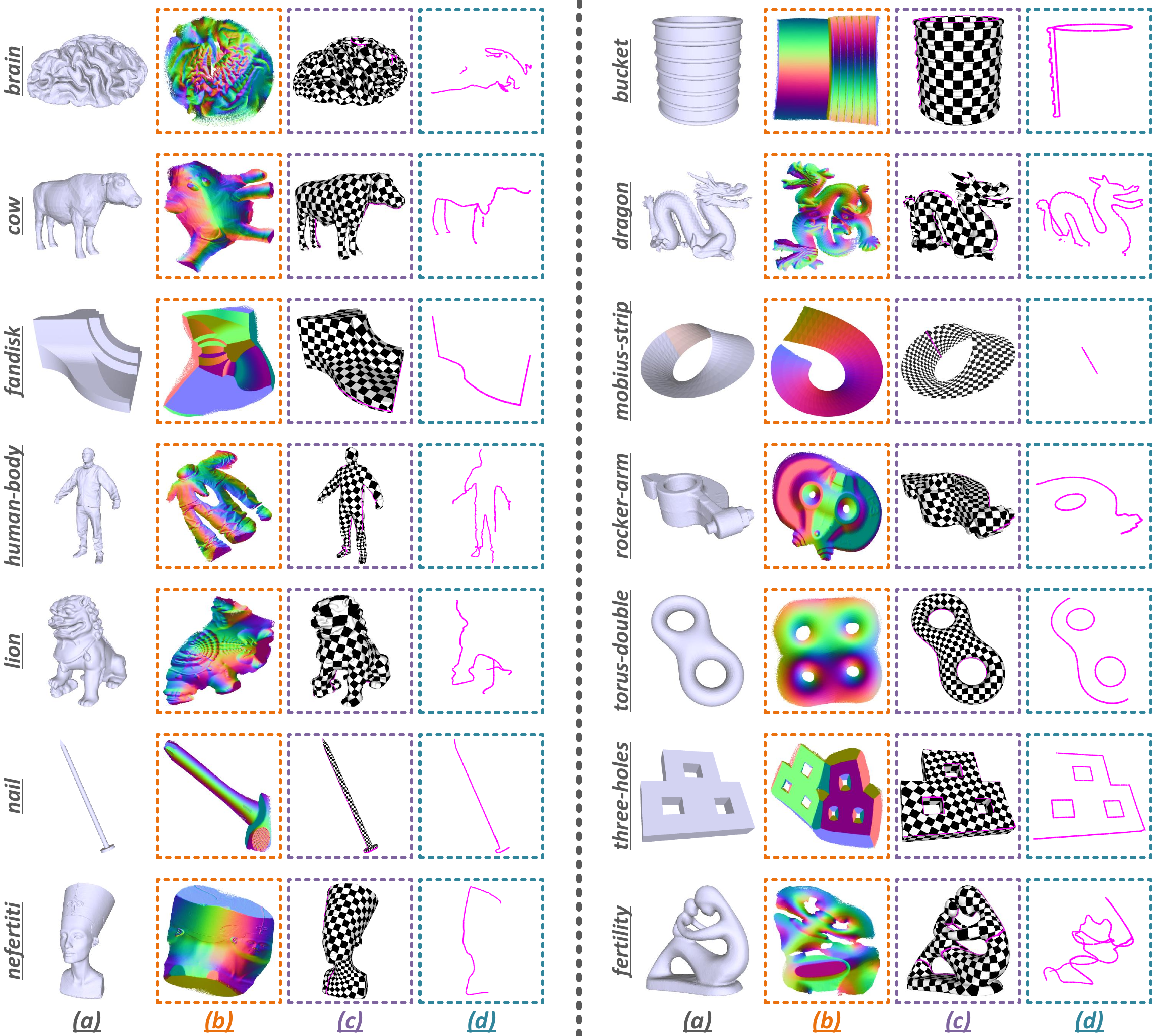}
	\caption{Display of surface parameterization results produced by our FAM. (a) input 3D models; (b) learned UV coordinates; (c) texture mappings; (d) learned cutting seams.}
	\label{fig:our_para}
	\vspace{-0.35cm}
\end{figure*}

\noindent \textbf{Parameterization with Different Geometric and Topological Complexities.} We conducted more comprehensive experimental evaluations, as displayed in Figure~\ref{fig:our_para} and Table~\ref{tab:quant_ours}, showing the universality and robustness of our approach.

\begin{figure*}[t!]
	\centering
	\includegraphics[width=0.90\linewidth]{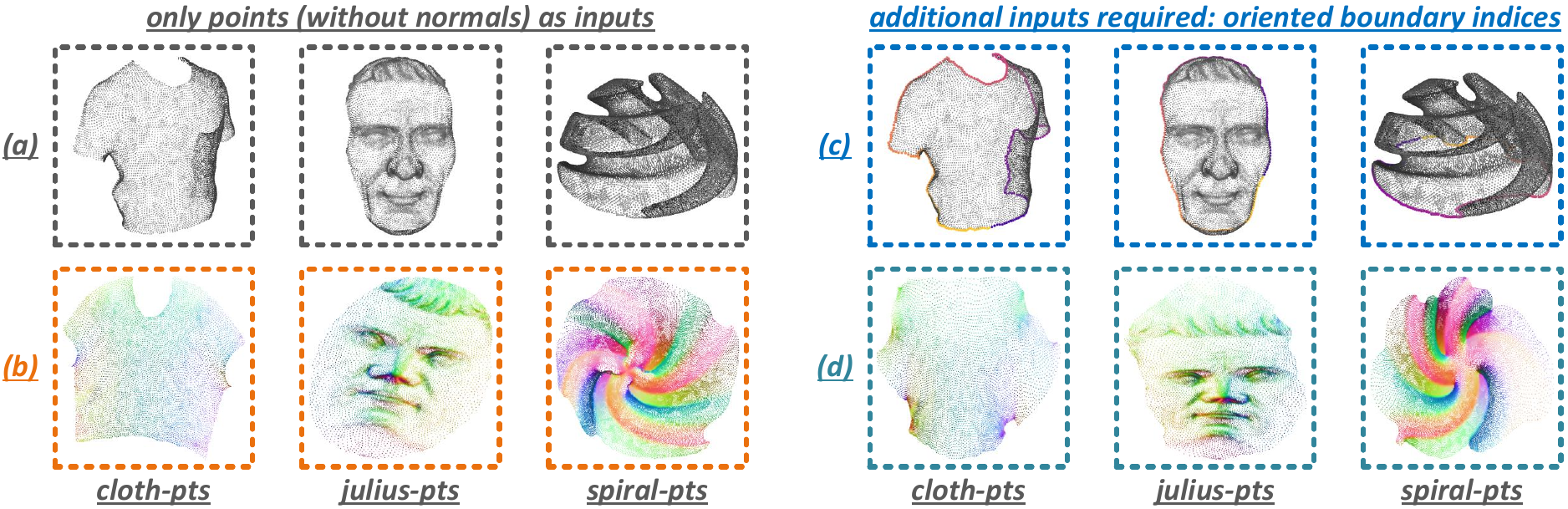}
	\caption{Point cloud parameterization achieved by our FAM (left) and FBCP-PC (right).}
	\label{fig:comparison_point_para}
	\vspace{-0.65cm}
\end{figure*}

\noindent \textbf{Comparison with FBCP-PC~\cite{choi2022free}.} Since FAM operates on discrete surface-sampled points without any dependence on connectivity information, applying our approach to parameterize unstructured and unoriented 3D point clouds is straightforward and basically seamless. The only modification is to remove the supervision of point-wise normals, i.e., the last term $\texttt{CS}(\mathbf{P}^\mathbf{n}; \mathbf{P}_\mathrm{cycle}^\mathbf{n})$ in Eqn.~(\ref{eqn:cc-loss}). As compared in Figure~\ref{fig:comparison_point_para} and Table~\ref{tab:quant_point_para_comparison}, our performances are not better than FBCP-PC. However, we must point out that FBCP-PC requires manually specifying indices of boundary points (arranged in order) as additional inputs. Hence, the comparisons are actually quite unfair to us.

\begin{figure*}[t!]
	\centering
	\includegraphics[width=0.95\linewidth]{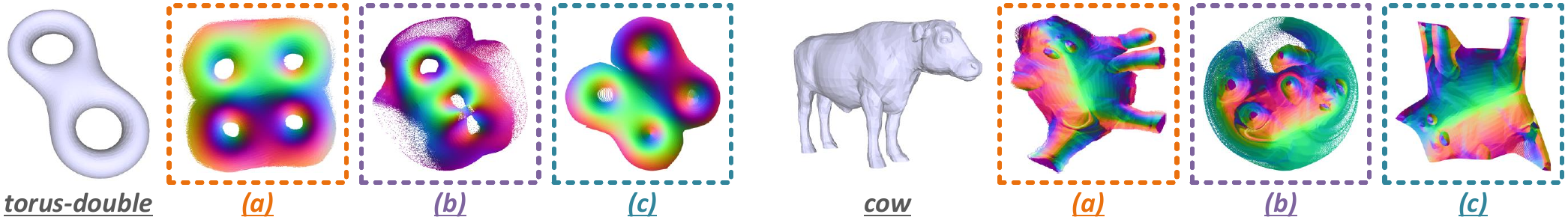}
	\caption{Ablation results produced by (a) our complete FAM framework, (b) FAM without the upper 2D$\rightarrow$3D$\rightarrow$2D learning branch, (c) FAM without the bottom 3D$\rightarrow$2D$\rightarrow$3D learning branch.}
	\label{fig:ablation_single_direction}
\end{figure*}

\begin{figure*}[t!]
	\centering
	\includegraphics[width=0.98\linewidth]{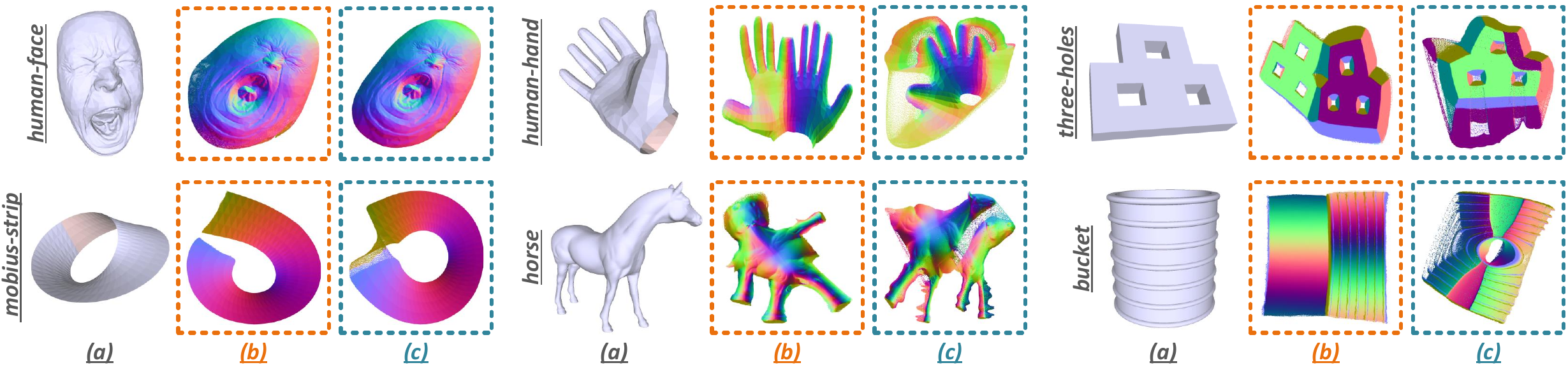}
	\caption{Ablation study on the necessity of Cut-Net: (a) input meshes; (b) results obtained from our full FAM architecture; (c) results obtained by removing the Cut-Net component.}
	\label{fig:ablation_cut-net}
\end{figure*}

\noindent \textbf{Ablation Studies.} We evaluated the necessity of our two-branch joint learning architecture. First, we removed the upper 2D$\rightarrow$3D$\rightarrow$2D branch to show the resulting $\mathbf{Q}$. Second, we removed the bottom 3D$\rightarrow$2D$\rightarrow$3D branch to show UV coordinates $\mathbf{\hat{Q}}$ by performing nearest-neighbor matching between $\mathbf{P}$ and $\mathbf{\hat{P}}$. As illustrated in Figure~\ref{fig:ablation_single_direction}, removing any of the two branches will cause different degrees of performance degradation. Furthermore, we verified the necessity of Cut-Net in the whole learning pipeline. As shown in Figure.~\ref{fig:ablation_cut-net}, removing Cut-Net does not lead to obvious performance degradation for models that are simpler to open (e.g., \textit{human-face} and \textit{mobius-strip}), yet for the other complex models the removal of Cut-Net causes highly-distorted surface flattening. Without learning offsets, the inherent smoothness of neural works can impede ``tearing'' the originally-continuous areas on the 3D surface, thus hindering the creation of cutting seams.

\begin{figure*}[t!]
	\makeatletter\def\@captype{table}\makeatother
	\begin{minipage}[b]{0.49\textwidth} 
		\centering	
		\renewcommand\arraystretch{2.1}
		\setlength{\tabcolsep}{6pt}
		\begin{tabular}{ c | c | c }
			\toprule[1.0pt]
			Testing Models & FAM & SLIM \\
			\hline\hline
			\makecell[c]{Open-Surface Models \\ (as in Figure~\ref{fig:comparison_with_slim})} & 0.156\% & 0.133\% \\
			\hline
			\makecell[c]{Higher-Genus Models \\ (as in Figure~\ref{fig:our_para})}  & 0.204\% & N/A     \\
			\bottomrule[1.0pt]
		\end{tabular}
		\vspace{-0.2cm}
		\caption{Quantitative self-intersection metrics of our parameterization results.}
		\label{tab:quant-self-intersect}
	\end{minipage}
	\makeatletter\def\@captype{figure}\makeatother
	\begin{minipage}[b]{0.49\textwidth} 
		\centering
		\includegraphics[width=1.0\textwidth]{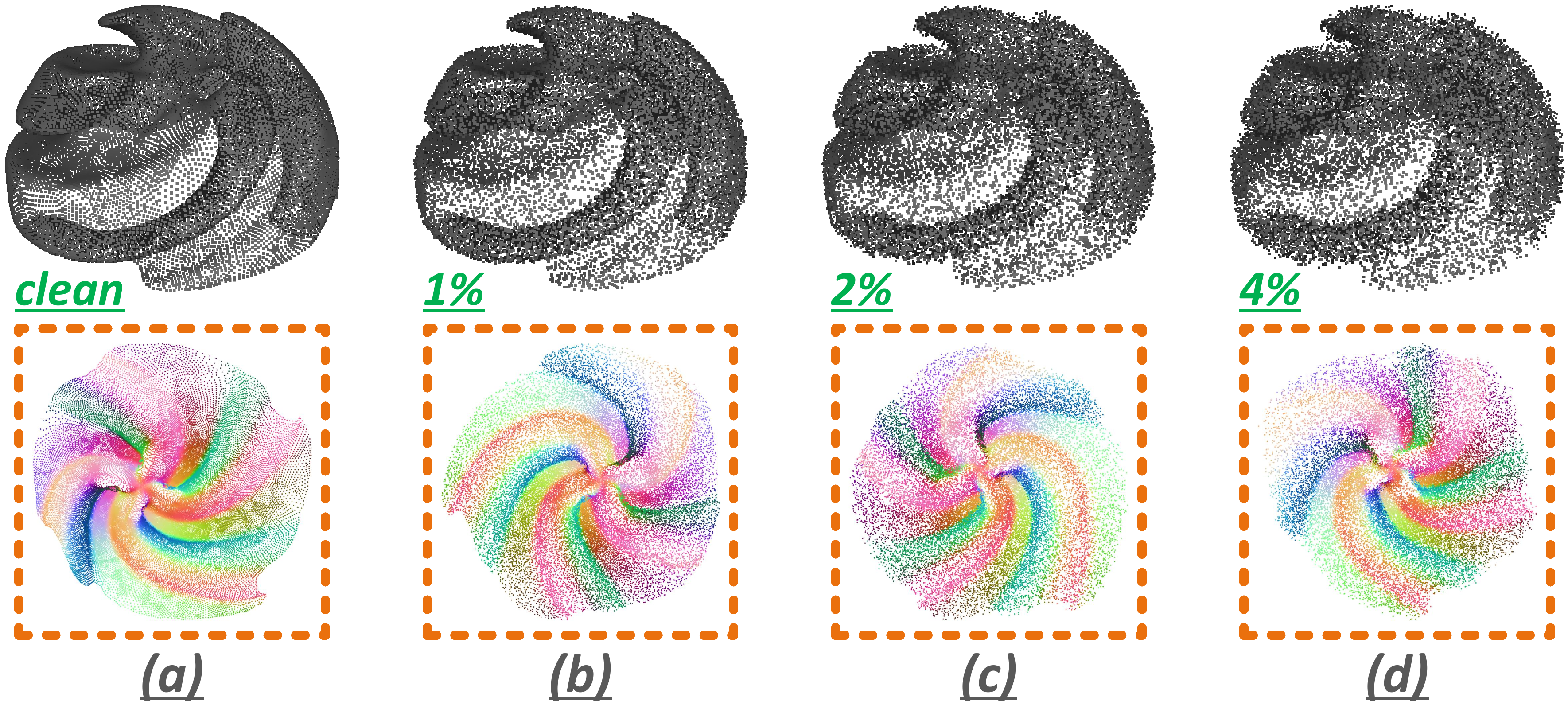} 
		\vspace{-0.65cm}
		\caption{Applying FAM to point clouds added with different levels of Gaussian \textbf{\textit{noises}}.}
		\label{fig:noisy-point-cloud}
	\end{minipage}
\end{figure*}

In addition, we conducted quantitative evaluations of self-intersection. Given a triangular mesh, we check if any pair of triangles overlaps in the UV space, and then measure the proportion of overlapped pairs to the total amount of triangle pairs. As reported in Table~\ref{tab:quant-self-intersect}, self-intersection are inevitable both in FAM and SLIM, but the proportions of self-intersected triangles are very small. In practice, it is not hard to apply post-processing refinement to slightly adjust the distribution of UV coordinates to further relieve or even eliminate self-intersection issues. For point cloud parameterization, we evaluated our robustness to noises in Figure~\ref{fig:noisy-point-cloud}, where we can observe that FAM shows stable performances to noisy input conditions (with 1\%, 2\%, and 4\% Gaussian noises for point position perturbation).

\begin{figure*}[t!]
	\centering
	\begin{subfigure}[b]{0.475\textwidth}
		\centering
		\includegraphics[width=\linewidth]{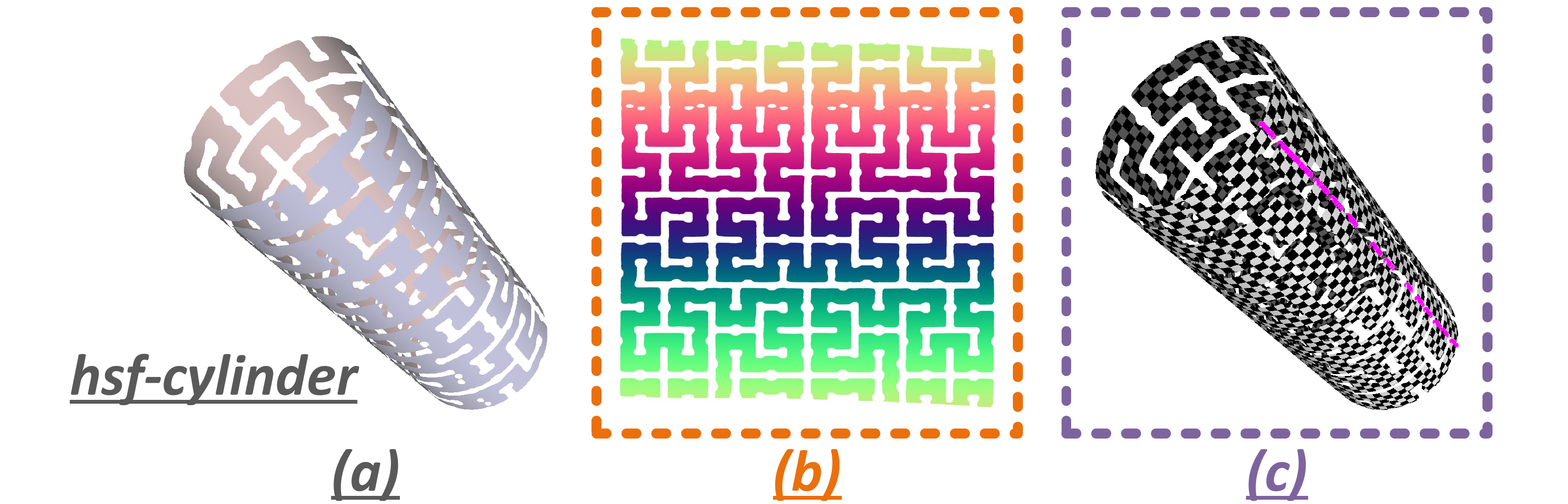}
		\caption{Highly challenging stress tests of our FAM.}
		\label{fig:stress-test-hilbert}
	\end{subfigure}\hfill
	\begin{subfigure}[b]{0.465\textwidth}
		\centering
		\includegraphics[width=\linewidth]{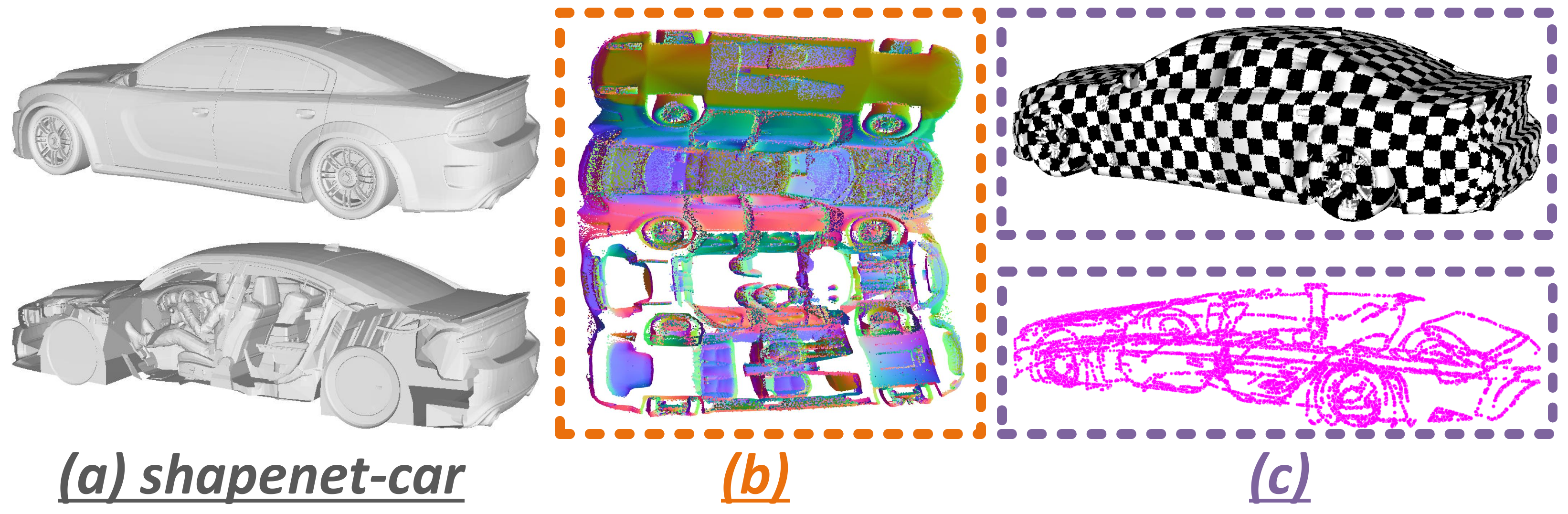}
		\caption{Failure cases of complicated CAD models.}
		\label{fig:stress-test-shapenet-cad}
	\end{subfigure}
	\vspace{-0.25cm}
	\caption{(a) input meshes; (b) UV maps; (c): texture mappings and learned cutting seams.}
	\vspace{-0.25cm}
\end{figure*}

Finally, we performed stress tests on a Hilbert-space-filling-shaped cylinder model in Figure~\ref{fig:stress-test-hilbert}. It is observed that our FAM obtains the basically optimal solution. Still, as shown in Figure~\ref{fig:stress-test-shapenet-cad}, processing the highly-complicated ShapeNet-style CAD model with rich interior structures and many multi-layer issues shows inferior UV unwrapping quality. Although our learned cutting seams are generally reasonable and texture mapping looks relatively regular from outside, it is hard to deal with the complicated interior structures.

\vspace{-0.1cm}
\section{Conclusion} \label{pp_sec:conclusion}
\vspace{-0.1cm}

We proposed FAM, the first neural surface parameterization approach targeted at global free-boundary parameterization. Our approach is universal for dealing with different geometric/topological complexities regardless of mesh triangulation quality and even applicable to unstructured point clouds. More importantly, our approach automatically learns appropriate cutting seams along the target 3D surface, and adaptively deforms the 2D UV parameter domain. As an unsupervised learning framework, our FAM shows significant potentials and practical values. The current technical implementations still have several aspects of limitations. Our working mode of per-model overfitting cannot exploit existing UV unwrapping results (despite their limited amounts) as ground-truths for training generalizable neural models. Besides, a series of more advanced properties, such as shape symmetry, cutting seam visibility, seamless parameterization, have not yet been considered.

\newpage

\appendix

\section{Appendix}

\subsection{Implementation Details}

In our bi-directional cycle mapping framework, all sub-networks are architecturally built upon stacked MLPs without batch normalization. We uniformly configured LeakyReLU non-linearities with the negative slope of $0.01$, except for the output layer. Within the Deform-Net, $\xi_d^{\prime}$ and $\xi_d^{\prime\prime}$ are four-layer MLPs with channels $[2, 512, 512, 512, 64]$ and $[66, 512, 512, 512, 2]$. Within the Wrap-Net, $\xi_w^{\prime}$ and $\xi_w^{\prime\prime}$ are four-layer MLPs with channels $[2, 512, 512, 512, 64]$ and $[66, 512, 512, 512, 6]$. Within the Cut-Net, $\xi_c^{\prime}$ and $\xi_c^{\prime\prime}$ are three-layer MLPs with channels $[3, 512, 512, 64]$ and $[67, 512, 512, 3]$. Within the Unwrap-Net, $\xi_u$ is implemented as three-layer MLPs with channels $[3, 512, 512, 2]$. In addition to network structures, there is also a set of hyperparameters to be configured and tuned. As presented in Eqn.~(\ref{eqn:find-cut}) for cutting seam extraction, we chose $K_\mathrm{cut} = 3$. Suppose that, at the current training iteration, the side length of the square bounding box of 2D UV coordinates $\mathbf{Q}$ is denoted as $L(\mathbf{Q})$. Then we set the threshold $T_\mathrm{cut}$ to be $2\%$ of $L(\mathbf{Q})$. Besides, as presented in Eqn.~(\ref{eqn:unwrap-loss}) for computing the unwrapping loss, we choose the threshold as $\epsilon = 0.2 \cdot L(\mathbf{Q}) / \sqrt{N}$, and $K_\mathrm{u}=8$. When formulating the overall training objective, the weights for $\ell_\mathrm{unwrap}$, $\ell_\mathrm{wrap}$, $\ell_\mathrm{cycle}$, and $\ell_\mathrm{conf}$ are set as $0.01$, $1.0$, $0.01$, and $0.01$. All our experiments are conducted on a single NVIDIA GeForce RTX 3090 GPU.

\begin{figure*}[h!]
	\centering
	\includegraphics[width=0.55\linewidth]{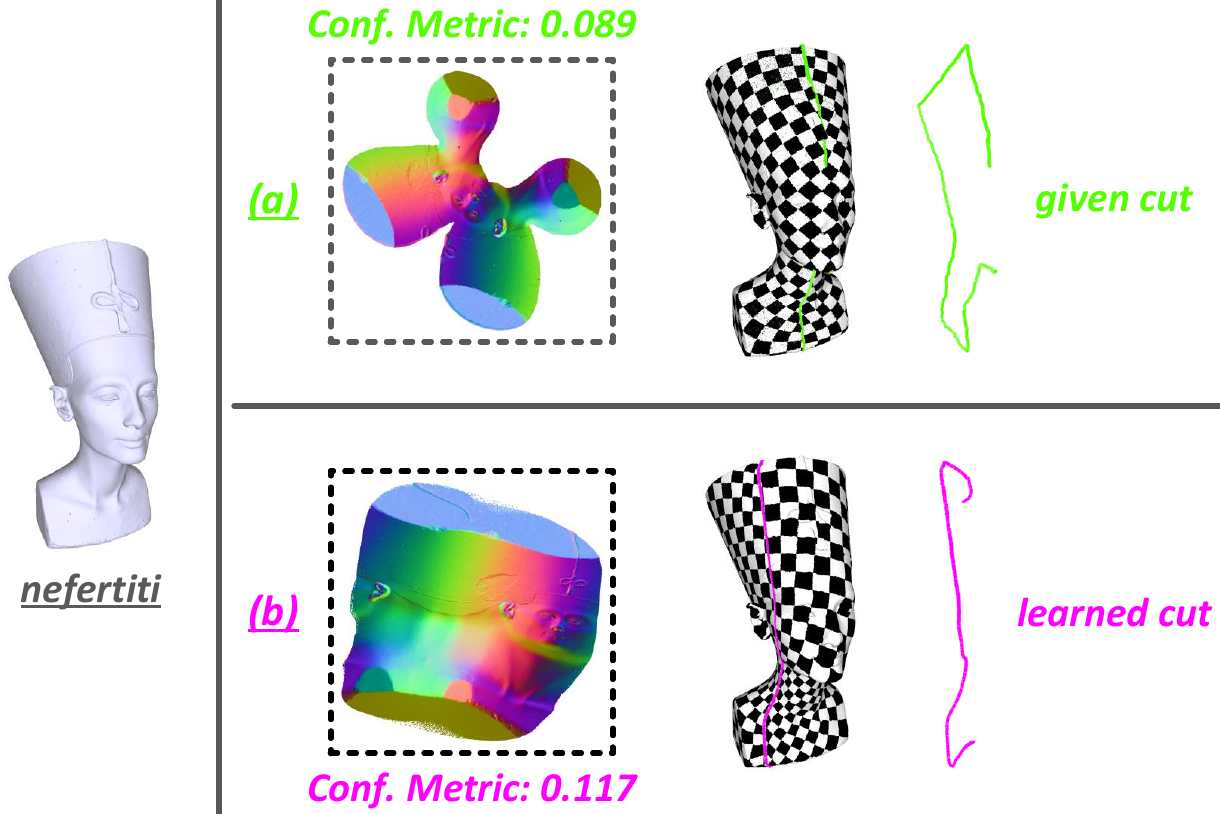}
	\caption{Experiments on the \textit{nefertiti} testing model whose topology is not homeomorphic to a disk. (a) The parameterization results of SLIM are obtained by \textbf{\textit{manually-specifying}} a high-quality cutting seam, with the conformality metric of $0.089$. (b) The parameterization results of our FAM, in which the cutting seam is automatically learned, with the conformality metric of $0.117$.}
	\label{suppl-fig:compare_results_given_cut}
	\vspace{-0.4cm}
\end{figure*}

\begin{table}[h!]
	\centering	
	\renewcommand\arraystretch{1.5}
	\setlength{\tabcolsep}{2.0pt}
	\caption{Optimization time costs (minutes) of our FAM and SLIM.}
	\begin{tabular}{ c | c c c c c c c c }
		\toprule[1.0pt]
		Model & \multicolumn{1}{c|}{\textit{human-face}} & \multicolumn{1}{c|}{\textit{human-head}} & \multicolumn{1}{c|}{\textit{car-shell}} & \multicolumn{1}{c|}{\textit{spiral}} & \multicolumn{1}{c|}{\textit{human-hand}} & \multicolumn{1}{c|}{\textit{shirt}} & \multicolumn{1}{c|}{\textit{three-men}} & \textit{camel-head} \\ 
		\hline
		SLIM  & \multicolumn{1}{c|}{39 min} & \multicolumn{1}{c|}{38 min} & \multicolumn{1}{c|}{19 min} & \multicolumn{1}{c|}{25 min} & \multicolumn{1}{c|}{28 min} & \multicolumn{1}{c|}{31 min} & \multicolumn{1}{c|}{17 min} & 40 min \\ 
		\hline
		\rowcolor{light-gray}
		FAM   & \multicolumn{8}{c}{around 18 min (basically unchanged for different models)} \\
		\bottomrule[1.0pt]
	\end{tabular}
	\label{suppl-tab:time-comparison-with-slim}
	\vspace{-0.2cm}
\end{table}

\subsection{Discussions about Global and Multi-Chart Surface Parameterization}

Over the years, global surface parameterization has continuously been the mainstream direction of research, since it potentially achieves smooth transitions and uniform distribution of mapping distortion across the entire surface, while multi-chart parameterization typically introduces discontinuities along patch boundaries, causing more obvious visual artifacts for texture mapping (perhaps the most important application scenario in the graphics field) and bringing additional difficulties in many other shape analysis tasks (e.g., remeshing, morphing, compression). For multi-chart parameterization, it is worth emphasizing that only obtaining chart-wise UV maps is not the complete workflow. We need to further perform chart packing to compose the multiple UV domains, without which the actual usefulness can be largely weakened. However, packing is known as a highly non-trivial problem, which is basically skipped in recent neural learning approaches, such as DiffSR and Nuvo. Still, local parameterization does have its suitable application scenarios for processing highly-complicated surfaces such as typical ShapeNet-style CAD meshes with rich interior structures and severe multi-layer issues, and per-chart distortion can be reduced since the geometry and topology of cropped surface patches become simpler.

\subsection{Comparison with SLIM on Non-Disk-Type Surfaces with Manually-Specified Cuts}

In the main manuscript, our experimental comparisons with SLIM focus on 3D models with open boundaries and/or disk topologies, on which SLIM can be applied. Here, we made additional efforts to conduct comparisons with SLIM on the non-disk-type 3D surface model of \textit{nefertiti} by manually specifying an optimal cutting seam for SLIM to run. Note that this is a quite \textbf{unfair} experimental setting, because finding optimal cuts is known to be a critical yet complicated task. As illustrated in Figure~\ref{suppl-fig:compare_results_given_cut}, our approach still achieves highly competitive performances compared with SLIM even under such an unfair comparison setup. 

\subsection{Running Efficiency}

In addition to parameterization quality, we further compared the time costs of our FAM and SLIM for training and optimization. Here, the official code of SLIM runs on the Intel(R) Core(TM) i7-9700 CPU. Table~\ref{suppl-tab:time-comparison-with-slim} lists the time costs for different testing models. Note that, since our FAM operates on surface-sampled points and we uniformly used the same number of input points (i.e., $N$), the time costs of our FAM basically maintain unchanged for different testing models. It turns out that our approach achieves satisfactory learning efficiency.

\begin{figure*}[t!]
	\centering
	\includegraphics[width=0.99\linewidth]{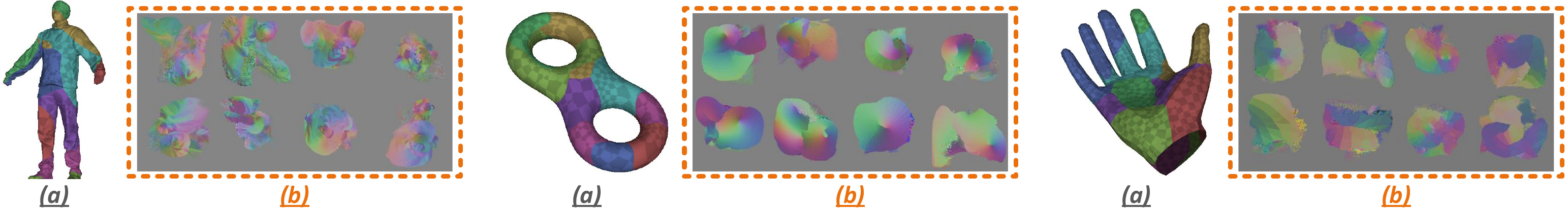}
	\caption{Results of Nuvo: (a) input meshes (different colors denote chart assignment); (b) chart-wise UV maps.}
	\label{fig:nuvo-results}
\end{figure*}

\begin{figure*}[t!]
	\centering
	\includegraphics[width=0.99\linewidth]{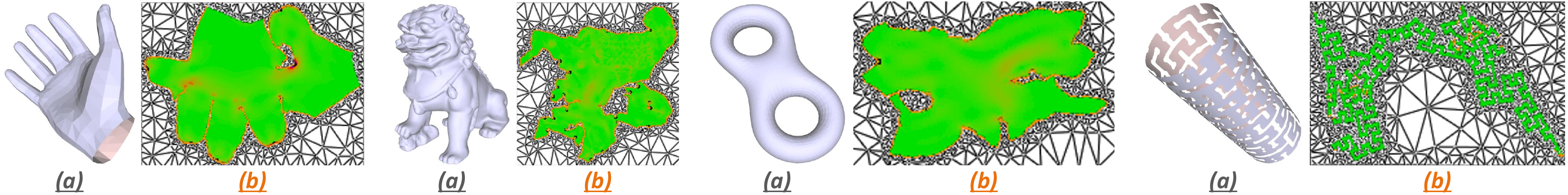}
	\caption{Results of OptCuts: (a) input meshes; (b) UV maps.}
	\label{fig:optcuts-results}
\end{figure*}

\subsection{Comparison with Nuvo~\cite{srinivasan2023nuvo}}

We provided typical parameterization results of Nuvo using a third-party implementation available at \url{https://github.com/ruiqixu37/Nuvo}, as presented in Figure~\ref{fig:nuvo-results}. We can observe that its chart assignment capability is still not stable. It often occurs that some spatially-disconnected surface patches are assigned to the same chart. Moreover, the critical procedure of chart packing is actually ignored in Nuvo. Directly merging the rectangular UV domains is not a valid packing. A real-sense packing should adaptively adjust the positions, poses, and sizes of each UV domain via combinations of translation, rotation, and scaling.

\subsection{Comparison with OptCuts~\cite{li2018optcuts}}

We provided typical parameterization results of OptCuts, as presented in Figure~\ref{fig:optcuts-results}. Comparatively, our neural parameterization paradigm still shows advantages in terms of flexibility (not limited to well-behaved meshes; applicable to point clouds), convenience (exploiting GPU parallelism; much easier for implementing and tuning), and parameterization smoothness (not limited to mesh vertices). Although OptCuts is able to jointly obtain reasonable surface cuts and UV unwrapping results, its performances are generally sub-optimal compared with ours.

\end{document}